\documentclass[twoside,11pt]{article}

\usepackage[abbrvbib, preprint]{jmlr2e}

\usepackage[title]{appendix}
\usepackage{adjustbox} 
\usepackage{multirow,booktabs} 
\usepackage{url}
\usepackage{subcaption}
\usepackage{color}

\usepackage{verbatim} 

\usepackage{bm}

\usepackage{lineno}
\usepackage{amsmath}

\usepackage{algorithm}
\usepackage[noend]{algorithmic}

\newcommand{\SetX}{{\mathcal{X}}}
\newcommand{\SetY}{{\mathcal{Y}}}
\newcommand{\SetR}{{\mathcal{R}}}
\newcommand{\SetT}{{\mathcal{T}}}
\newcommand{\Vecx}{{\bm{x}}}
\newcommand{\Vecy}{{\bm{y}}}
\newcommand{\Vecr}{{\bm{r}}}
\newcommand{\Vect}{{\bm{t}}}
\newcommand{\Vecdelta}{{\bm{\delta}}}

\newcommand{\MatB}{{\bm{B}}}
\newcommand{\MatD}{{\bm{D}}}

\newcommand{\DefR}{{\mathbb{R}}}

\DeclareMathOperator*{\argmin}{argmin}

\newcommand{\Vecd}{{\bm{d}}}
\newcommand{\Vecb}{{\bm{b}}}
\newcommand{\MatA}{{\bm{A}}}

\jmlrheading{x}{xxxx}{x-xx}{xx/xx}{xx/xx}{meila00a}{Joonas {H{\"a}m{\"a}l{\"a}inen}, Alisson S. C. Alencar, Tommi {K{\"a}rkk{\"a}inen}, C\'{e}sar L. C. Mattos, Amauri H. Souza J\'{u}nior, and Jo\~{a}o P. P. Gomes}

\ShortHeadings{MLM: Theoretical Results and Clustering-Based Reference Point Selection}{{H{\"a}m{\"a}l{\"a}inen}, Alencar, {K{\"a}rkk{\"a}inen}, Mattos, Souza J\'{u}nior, and Gomes}
\firstpageno{1}

\begin{document}

\title{Minimal Learning Machine: Theoretical Results and Clustering-Based Reference Point Selection}

\author{\name Joonas H{\"a}m{\"a}l{\"a}inen \email joonas.k.hamalainen@jyu.fi \\
       \addr University of Jyvaskyla, Faculty of Information Technology\\
		     P.O. Box 35, FI-40014 University of Jyvaskyla, Finland
       \AND
       \name Alisson S. C. Alencar \email alencar.alisson@lia.ufc.br\\
       \addr Federal University of Cear\'{a} - UFC, Department of Computer Science\\
			Fortaleza-CE, Brazil
       \AND
       \name Tommi K{\"a}rkk{\"a}inen \email tommi.karkkainen@jyu.fi \\
       \addr University of Jyvaskyla, Faculty of Information Technology\\
		     P.O. Box 35, FI-40014 University of Jyvaskyla, Finland
       \AND
       \name C\'{e}sar L. C. Mattos \email cesarlincoln@dc.ufc.br\\
       \addr Federal University of Cear\'{a} - UFC, Department of Computer Science\\
			Fortaleza-CE, Brazil
       \AND
       \name Amauri H. Souza J\'{u}nior \email amauriholanda@ifce.edu.br\\
       \addr Federal Institute of Education, Science and Technology of Cear\'{a} - IFCE\\
		     Department of Computer Science, Maracana\'{u}-CE, Brazil
       \AND
       \name	 Jo\~{a}o P. P. Gomes \email jpaulo@dc.ufc.br\\
       \addr Federal University of Cear\'{a} - UFC, Department of Computer Science\\
			Fortaleza-CE, Brazil}

\editor{xxxx xxxx}

\maketitle

\begin{abstract}
The Minimal Learning Machine (MLM) is a nonlinear supervised approach based on learning a linear mapping between distance matrices computed in the input and output data spaces, where distances are calculated using a subset of points called reference points. Its simple formulation has attracted several recent works on extensions and applications. In this paper, we aim to address some open questions related to the MLM. First, we detail theoretical aspects that assure the interpolation and universal approximation capabilities of the MLM, which were previously only empirically verified. Second, we identify the task of selecting reference points as having major importance for the MLM’s generalization capability. Several clustering-based methods for reference point selection in regression scenarios are then proposed and analyzed. Based on an extensive empirical evaluation, we conclude that the evaluated methods are both scalable and useful. Specifically, for a small number of reference points, the clustering-based methods outperformed the standard random selection of the original MLM formulation.
\end{abstract}


\section{Introduction}
\label{sec:intro}

Machine learning techniques can be roughly categorized as unsupervised and supervised, depending on whether the learning data comprises only input data or a complete set of input-output pairs \citep{shalev2014understanding}. In terms of target data, semi-supervised learning typically lies somewhere in the middle of these extremes \citep{gan2013using}, and active \citep{aggarwal2014active} or incremental \citep{losing2018incremental} learning techniques acquire the desired outputs during model construction incrementally, on a need-to-know basis. A key concept in unsupervised learning, especially clustering, is the distance or dissimilarity between two observations or for an observation-metaobservation (e.g., cluster prototype) pair \citep{reddy2013survey}. Currently, supervised learning is extensively using deep structures of multiple layers of weights and stochastic optimization in training \citep{hubara2017quantized}.

The distance-based supervised methods provide a methodological middle ground and linkage between unsupervised and supervised learning. 
Recent examples of such methods include the Minimal Learning Machine \citep{de2015minimal} and the Extreme Minimal Learning Machine \citep{KarNeCo2019}. The core learning construct in these methods is the distance regression, based on the dissimilarity between the observations. Hence, nonlinear regression and classification can be performed for all entities where the dissimilarity can be metrically defined. During learning, incremental use of the so-called reference points, together with the solution of the corresponding distance-based linear system, is needed without any optimization procedure \citep{KarNeCo2019}. Note that such distance-based supervised techniques also enable direct utilization of metric learning techniques as part of their construction (e.g., \citealt{MAL-019}).

The increasing popularity of the MLM can be explained by its simple formulation, easy implementation, and promising results in several applications \citep{mesquita2017ensemble,coelho2014,marinho2017novel,marinho2018novel,pihlajamaki2020monte}. Apart from the applications of the MLM, many studies have been carried out in recent years to improve and augment the basic form of the MLM to handle missing values \citep{mesquita2015minimal, mesquita2017euclidean} and outliers \citep{RobustMLMesann2017}, perform ensemble learning \citep{mesquita2017ensemble} and semi-supervised learning \citep{caldas2018fast}, speed up its computations \citep{florencio2020new, mesquita2017ensemble, marinho2016new}, and include a reject option in classification tasks \citep{de2016efficient}.

\subsection{Prior Work on Distance-based Learning}
\label{subsec:priorwork}


Radial basis function networks (RBFN) \citep{Powell1987,BroLowComSys1988} popularized the use of distance to the training data as part of  neural network models. Usually the distance in RBFN is further transformed with a nonlinear activation function, but early papers analyzing the technique explicated also the use of a linear, distance-based kernel \citep{poggio1990networks,park1991universal}. 

The more actual development of dissimilarity-based machine learning techniques was advanced by \cite{pekalska2001automatic}, who proposed to use ``global classifier defined on the similarities to a small set of prototypes, called the representation set.'' This representation set is the set of reference points in the MLM lingo. Moreover, similarly to Step 1 in MLM (see Section \ref{mlm}), dissimilarities and the corresponding distance matrix between the objects in the representation and training sets were used in \citep{pekalska2001automatic}, where the regularized linear normal density classifier was applied. A linear classifier model based on dissimilarities was then proposed in \cite{pekalska2001generalized}, with parameters estimated similarly to the SVM by solving a linear programming problem for the separating hyperplane in binary classification. Fisher linear discriminant was used for the distance-based spectral classification in \citep{paclik2003dissimilarity}.

According to \cite{balcan2008theory}, use of the euclidean distance function corresponds to the trivial, identity kernel and to the corresponding scaled similarity function \citep[Definition 1]{balcan2008theory}. Let us refer this as the euclidean kernel in what follows. This means that the distance transformation in the MLM introduces the famous kernel trick, where the size of the implicit space coincides with the number of the reference points. However, because the whole construction of the kernelized learning in MLM happens in the distance space, this formulation is different from the SVM or kernel-perceptron, or from the previously proposed approaches with dissimilarity kernels in \citep{pekalska2001automatic,pekalska2001generalized,
paclik2003dissimilarity,pkekalska2006prototype,
pekalska2008beyond,chen2009similarity,wang2009theory}.

Closely related work to ours, again in the context of Step 1 of the MLM, is \citep{zerzucha2012dissimilarity}, where the complete euclidean dissimilarity matrix is used with the partial least squares method. Fuzzy clustering and leave-one-out cross-validation are suggested for the identification of the most informative subset of data (i.e., reference point selection) for the reduced euclidean distance matrix.

Feature selection combined with distance-based classification of imbalanced data was considered in \cite{zhang2015dissimilarity}, where
Naive Bayes, instance-based nearest neighbor, Random Forest, Multilayer Perceptron, and Logistic Regression from WEKA were used as classifiers. Note that SVM is the dominant (practically only) method that has been used with the distance-based kernels for the time series classification \citep{abanda2019review}. Dissimilarity-based method with random forest as classifier was proposed in \citep{cao2019random}. A recent review on various dissimilarity-based approaches is given in \citep{costa2020dissimilarity}.

%
%

In conclusion, the use of distances or dissimilarities or proximities in supervised learning is not new  (e.g.,  \cite{balcan2008theory,chen2009similarity,
schleif2015indefinite}). As summarized by  \cite{chen2010strategies}, the most straightforward utilization of distance calculations is to use the pairwise distances as features of a predictive model. Indeed, this is part of the MLM, which is additionally characterized by reference point selection, genuine distance regression, and solution of a multilateration problem. Hence, the whole learning framework with MLM is different from earlier work in the field as depicted, e.g., in \cite{pekalska2001automatic,paclik2003dissimilarity,
pkekalska2006prototype,wang2007learning,
pekalska2008beyond,balcan2008theory,nguema2008model,
chen2009similarity,chen2010strategies,
schleif2015indefinite}.

\subsection{Importance of Reference Point Selection}
\label{subsec:refpointexample}

In the MLM, reference points are a subset of the training points and are used to build the distance matrices that are a key part of the MLM’s induction process. In the original MLM formulation, the reference points were randomly selected. As empirically demonstrated by \cite{de2015minimal}, a bad choice of reference points can damage the MLM’s generalization capability. This phenomenon is even more likely to occur when the number of reference points is small \citep{de2015minimal}. An illustrative example on the effects of different reference point selection strategies is given in Section \ref{subsec:RefPointMot}.

There exists also a large number of proposals to improve the behaviour of the distance-based methods using reference point (also referred as prototypes or landmarks) selection.
\cite{pkekalska2006prototype}
considered prototype/reference point selection with Bayesian classifiers. They concluded
that a set of few, evenly distributed centers provided a better classification results (higher accuracy faster) than the use of all training examples.
Later suggestions towards this direction were given in \citep{plasencia2014towards,plasencia2017scalable}, again in the form of finding a small set of prototypes. A well-spread set of diverse reference points 
was also suggested as part of the similarity based learning framework in \cite{kar2011similarity}.

\cite{RSesann2018} proposed a strategy to select reference points based on the identification of the class boundaries in a binary classification problem. In the proposal, it was prohibited to select any point as a reference point from a subset of points in the class boundary area. A similar objective was pursued by \cite{florencio2018}, who identified such a region using fuzzy c-means. \cite{Maia2018} used a sparse regression method to build the linear mapping between distance matrices. In that work, the reference points were selected according to the resulting non-zero coefficients obtained by the linear model.

Even if the previous work on reference point selection led to more compact models with better generalization, the existing efforts only focused on classification problems. Additionally, none of these works presented any theoretical results that could explain the impact of choosing reference points in a general setting.

\subsection{Contributions}
\label{subsec:contrib}

In the present work, we advance the research field as depicted above as follows: i) by presenting a proof of the MLM’s interpolation capability when all training points are used as reference points; ii) by demonstrating the universal approximation capability of the MLM even in scenarios in which reference point selection is considered; and iii) by proposing and analyzing several reference point selection strategies for regression problems based on elements of clustering methods.

When we choose clustering-based approaches, our basic hypothesis is that a set of well-spread reference points in the data space will improve the performance of the MLM compared to random selection. We validate the empirical contributions of this paper through computational experiments with 15 regression datasets.

The remainder of the paper is organized as follows: 
Section \ref{mlm} presents the basic formulation of the MLM. Section \ref{theory} details our theoretical contributions on the interpolation and generalization capabilities of the MLM. Section \ref{prop} describes clustering-based methodologies of reference point selection. Section \ref{exp} presents a comprehensive set of experiments to evaluate clustering-based methodologies for reference point selection. Finally, Section \ref{concl} concludes the paper.

\section{Minimal Learning Machine} 
\label{mlm}

As previously discussed, the MLM is a distance-based supervised machine learning method. The basic algorithm \citep{de2013minimal,de2015minimal} comprises two main steps: i) regression estimation using the distance-based kernel; and ii) distance-based interpolation of a new output. For clarity, we describe these two steps below.

Let $\SetX = \{ \Vecx_i \}_{i=1}^{N}$ be a set of training inputs, where $\Vecx_i \in \DefR^P$, and $\SetY = \{ \Vecy_i \}_{i=1}^{N}$ is the set of the corresponding outputs, for $\Vecy_i \in \DefR^L$, respectively. Moreover, we define the set of (input) reference points $\SetR = \{ \Vecr_k \}_{k=1}^{K}$ as a non-empty subset of $\SetX$, $\SetR \subseteq \SetX$, and let $\SetT = \{ \Vect_k \}_{k=1}^{K}$ refer to the outputs of the corresponding reference inputs, i.e., $\Vecr_k \mapsto \Vect_k$.

Next, we define two distance matrices, $\MatD_x \in \DefR^{N\times K}$ and $\MatD_y \in \DefR^{N\times K}$, using the Euclidean distance $\| \cdot \|$ as follows:
\begin{equation}\label{Dx}
\MatD_x = \begin{bmatrix} \| \Vecx_i - \Vecr_k \| \end{bmatrix} \quad i = 1, \ldots, N,\ k = 1, \ldots, K,
\end{equation}
\begin{equation}\label{Dy}
\MatD_y = \begin{bmatrix} \| \Vecy_i - \Vect_k \| \end{bmatrix} \quad i = 1, \ldots, N,\ k = 1, \ldots, K .
\end{equation}
The key idea for the first step of the MLM is the assumption of a regression model between the distance matrices: $\MatD_y = g (\MatD_x) + \bm{E}$, where $\bm{E}$ denotes the residuals/error in this transformation. Assuming that the unknown regression model is of the linear form, its transformation matrix $\MatB \in \DefR^{K\times K}$ can be estimated using the well-known ordinary least squares formulation, as follows:
\begin{equation}\label{B}
\MatB = \left( \MatD_x^T \MatD_x \right)^{-1} \MatD_x^T \MatD_y .
\end{equation}
The linear mapping represented by the matrix $\MatB$, obtained in Eq. \eqref{B}, is the first step of the MLM.

For the second step, let $\tilde{\Vecx}$ be a new input vector whose output needs to be estimated. Hence, based on the distance regression model from the first step, we seek the corresponding output $\tilde{\Vecy}$, satisfying
\begin{equation}\label{Yestim}
\| \tilde{\Vecy} - \Vect_k \| \approx \delta_k \quad \forall k = 1, \ldots, K,
\end{equation}
where
$$
\Vecdelta = \begin{bmatrix} \| \tilde{\Vecx} - \Vecr_k \| \end{bmatrix}_{k=1}^K \MatB .
$$
The solution to the multilateration problem in Eq. \eqref{Yestim} can also be obtained using the least-squares formulation by letting
\begin{equation}\label{Yestim2}
\tilde{\Vecy}^* = \argmin \mathcal{J} (\tilde{\Vecy}), \quad
\text{where } \ \mathcal{J} (\tilde{\Vecy}) = \sum_{k=1}^K \left( \| \tilde{\Vecy} - \Vect_k \|^2 - \Vecdelta_k^2 \right)^2 .
\end{equation}

As stated by \cite{de2015minimal}, there are many possible solvers for Eq. \eqref{Yestim2}. In the original formulation, the MLM solves the output estimation step by using a nonlinear optimization algorithm. Such an algorithm is used to find the point that minimizes the double-quadratic error between the estimated distance and the real distance, calculated on each candidate point. However, we want to verify whether, when the distances are perfectly estimated, the position of the point can be recovered without error. To that end, we follow an alternative formulation of the multilateration problem, called the localization linear system (LLS), detailed in Appendix \ref{ap:LLS}. This formulation provides an efficient method for the output estimation. The LLS method computes the output position by solving a linear system. An output prediction algorithm for the MLM with LLS is depicted in the Algorithm \ref{alg:OutPredLLS}. Substitution ``$\gets \small[ ~ \small]$'' referes to the removal of an element from a vector.

\begin{algorithm}[!t]
\caption{MLM output prediction with LLS}
\label{alg:OutPredLLS}
\begin{algorithmic}[1]
\REQUIRE input $\tilde{\Vecx}$, distance regression model $\MatB$, reference points $\SetR$ and $\SetT$.
\ENSURE predicted output $\tilde{\Vecy}$.
	\STATE $\Vecd_{\tilde{\Vecx}} \gets \begin{bmatrix} \| \tilde{\Vecx} - \Vecr_k \| \end{bmatrix}_{k=1}^K$
    \STATE $\Vecdelta \gets \Vecd_{\tilde{\Vecx}} \MatB$ ~~~~~~~~~~~~~~~~~~~~~~~~~~~~~~~~// \text{Predict distances in output space}
    \STATE $i^* \gets rand(\{ 1, \ldots, K \})$
    \STATE $\Vect^* , \delta^* \gets \Vect_{i^*}, \Vecdelta(i^*)$
	\STATE $\SetT , \Vecdelta(i^*), K \gets \SetT \backslash \{ \Vect_{i^*} \} ,  \small[ ~ \small] , K - 1 $ ~~// \text{Remove BAN from the set of reference points} 
	\STATE $\Vecb, \MatA$
	\FOR {$i \in \{  1, \ldots, K \}$} 
	\STATE $\Vecb(i) \gets \frac{1}{2}({\delta^*}^2 + {\| \Vect^* - \Vect_i \|}^2  -  {\Vecd_{\tilde{\Vecx}}(i)}^2)$
	\STATE $\MatA(i,:) \gets  ({\Vect_i} - {\Vect^*})^T$
	\ENDFOR
	\STATE $\bm{\theta} \gets solve( \MatA \bm{\theta} = \Vecb )$ ~~~~~~~~~~~~~~~~~~// \text{Solve a linear system of equations} 
	\STATE $\tilde{\Vecy} \gets \bm{\theta} + \Vect^*$
 \end{algorithmic}
\end{algorithm}

In summary, the LLS solves a system in the form $\bm{A \theta} = \bm{b}$. The coefficient matrix $\bm{A}$ is constructed based on all but one reference point, named \textit{benchmark-anchor-node} (BAN), and each row $i$ is given by the difference between the $i$-th reference point and the BAN. The vector $\bm{\theta}$ is a simple translation of the target position. The vector $\bm{b}$ is computed from the estimated distances between the target point and the reference points, as well as the distance from the BAN itself to the other reference points.

In Algorithm \ref{alg:OutPredLLS}, a linear system of equations in Step 10 is usually overdetermined. An approximate solution can be obtained from the ordinary least squares (OLS) method with a computational cost of $\mathcal{O}(L^2K)$. Step 1 has a computational cost of $\mathcal{O}(KP)$. Usually, Step 2 is computationally the most expensive step and determines the asymptotic behavior of the computational complexity, $\mathcal{O}(K^2)$, when $K >> P$ and $K >> L$. Therefore, models with a reduced number of reference points can lead to significant computational time reduction of the MLM prediction with the LLS when the input and output space dimensions are small with respect to $K$.


%

\paragraph{Nystr{\"o}m approximation}
Initially, one could draw similarities between the MLM formulation and the methods that consider a Nystr{\"o}m approximation for Gram matrices \citep{williams2001using, drineas2005nystrom, sun2015review}. Such methods are based on the approximation $\bm{K} \approx \bm{C} \bm{W}^\dagger \bm{C}^T$, where $\bm{K} \in \mathbb{R}^{N \times N}$, $\bm{W} \in \mathbb{R}^{K \times K}$, $\bm{C} \in \mathbb{R}^{N \times K}$, and $K \ll N$.

However, the exact least squares solution $\bm{B} = (\bm{D}_x^T \bm{D}_x)^{-1}\bm{D}_x^T \bm{D}_y$ in the first step of the MLM is different. In fact, for $K < N$ the distance matrices $\bm{D}_x \in \mathbb{R}^{N \times K}$ and $\bm{D}_y \in \mathbb{R}^{N \times K}$ are rectangular and we cannot obtain the same solution by directly applying the standard Nystr{\"o}m approach. We confirm the latter statement by considering the full distance matrices $\bm{\Delta}_x \in \mathbb{R}^{N \times N}$ and $\bm{\Delta}_y \in \mathbb{R}^{N \times N}$, which correspond to the solution $\bm{B} = \bm{\Delta}_x^{-1} \bm{\Delta}_y$. By considering a Nystr{\"o}m representation for $\bm{\Delta}_x$ we would obtain $\bm{B} \approx (\bm{C} \bm{W}^\dagger \bm{C}^T)^{-1} \bm{\Delta}_y$. The inverse $(\bm{C} \bm{W}^\dagger \bm{C}^T)^{-1}$ may exist only for $K=N$, the only case where we recover the least squares solution by choosing, for instance, $\bm{C} = \bm{W} = \bm{\Delta}_x$.

Nevertheless, the vectors used to build the matrix $\bm{W}$ in a Nystr{\"o}m approximation, usually called \textit{landmarks}, can be seen as analogous to the reference points in the MLM. In the Nystr{\"o}m method literature it is well known that clustering algorithms are a sensible strategy for choosing landmarks \cite{zhang2008improved, zhang2010clustered, kumar2012sampling, oglic2017nystrom, pourkamali2018randomized}. Such an observation encourages us to also pursue a clustering approach for selecting the MLM reference points.

\section{MLM Theoretical Results}
\label{theory}

In this section, we detail some theoretical guarantees of the MLM. These results are divided into two subsections: interpolation theory and universal approximation capability.

\subsection{Interpolation Theory}

We show that the MLM can interpolate the data in two steps. First, we show that the distance matrix $D_x$, constructed using all points in the available data as reference points, is invertible. According to Eq. \eqref{B} and given that $D_x^T = D_x$ when
all data points act as reference points, the distances can be estimated accurately. In the second step, we prove that, under certain conditions to be described, the estimation of the output will recover the position of the original points with zero error.
        
\subsubsection{Inverse of distance matrices}
		
In the training phase of the MLM, we need to solve a linear system whose coefficient matrix is given by the distances between the points of the dataset and the reference points, that is, a matrix $D_x$ such that $d_{i, j}$ is given by $d(\bm{x_i}, \bm{r_j})$, i.e., the distance between the $i$-th point of the training set and the $j$-th reference point. If we consider the specific case in which all points in the dataset were reference points, then the coefficient matrix was a square matrix of order equal to the number of training points $N$. We rearrange the points so that $x_i = r_i, \forall i \in \{1, \cdots, N\}$, and the matrix of coefficients is such that each element $d_{i, j}$ is given by $d(\bm{x}_i, \bm{x}_j)$. A matrix with this characteristic is formally called a \textbf{distance matrix}. To find an exact solution, we must show that every distance matrix admits an inverse.

The invertibility of the distance matrix was first demonstrated by \cite{Micchelli1986}; \cite{Auer95} offered a simplified proof. The main result is given by the following theorem:

        \begin{theorem}
        Given a distance matrix $\bm{D}$ computed from a set of $N$ distinct points, the determinant of $\bm{D}$ is positive if $N$ is odd and negative if $N$ is even; specifically, $\bm{D}$ is invertible.
        \end{theorem}         
 
With this result, we can guarantee that when the distance matrix in the input space is multiplied by the coefficient matrix obtained in the MLM training, the result is the distance matrix in the output space, without any error.
		
\subsubsection{Condition for the perfect estimation of the multilateration}  
\label{multSolver}

The result of the previous subsection is important since it shows that the MLM can recover the distances in the output space between the reference points and the training data with zero errors. However, this is not sufficient evidence to say the MLM is capable of interpolating any dataset. For that, we must prove the model’s ability to estimate the output, i.e., to retrieve the position of the points in the output space from the perfectly estimated distances.
	    	
%
%

Solving the LLS accurately is only possible when the coefficient matrix is non-singular. This is not necessarily true for any set of points. In fact, the theorem below shows that the matrix is invertible when the reference points, including the BAN, form an independent affine set.

\begin{theorem}[Perfect estimation with the multilateration]
	Given a linearly independent spanning set $\bm{v}_1\hdots \bm{v}_R \in \mathbb{R}^S$ and $\bm{v} \in \mathbb{R}^S$. If $\bm{v}$ is not an affine combination of $\{\bm{v}_1,\hdots,\bm{v}_R\}$, then the set of vectors $\bm{v}_1 - \bm{v}, \bm{v}_2 - \bm{v},\hdots,\bm{v}_R - \bm{v}$ is linearly independent.
\end{theorem}       

\noindent
\textbf{Proof:}

	Suppose that $\bm{V} = \{\bm{v}_1,\hdots,\bm{v}_R\}$ is a linearly independent spanning set, $\bm{v}$ is not an affine combination of $\bm{V}$, and $\bm{V}' = \{\bm{v}_1 - \bm{v},\hdots,\bm{v}_R - \bm{v}\}$ is linearly dependent. There then exists $\mu_1,\hdots,\mu_R$, not all equal to zero, such that
    \begin{eqnarray}
		\sum_{i = 1}^R \mu_i(\bm{v}_i - \bm{v}) = 0 \nonumber \\
        \sum_{i = 1}^R \mu_i(\bm{v}_i - \sum_{j = 1}^R \lambda_j \bm{v}_j ) = 0 \nonumber \\
        \sum_{i = 1}^R \mu_i\bm{v}_i - \sum_{i = 1}^R \mu_i\sum_{j = 1}^R \lambda_j \bm{v}_j = 0 \nonumber \\
        \sum_{i = 1}^R \mu_i\bm{v}_i - \sum_{i = 1}^R \lambda_i \bm{v}_i\sum_{j = 1}^R \mu_j = 0 \nonumber \\
        \sum_{i = 1}^R (\mu_i\bm{v}_i - \lambda_i \bm{v}_i\sum_{j = 1}^R \mu_j) = 0 \nonumber \\
        \sum_{i = 1}^R \underbrace{(\mu_i - \lambda_i\sum_{j = 1}^R \mu_j)}_{\theta_i}\bm{v}_i = 0 \nonumber \\
        \label{eq:mult_proof}
        \sum_{i = 1}^R \theta_i\bm{v}_i = 0.
	\end{eqnarray}
    Since $\bm{V}$ is LI, Eq. \eqref{eq:mult_proof} can only be satisfied when all $\theta_i$ are equal to zero, which means that $\mu_i = \lambda_i \sum \mu_j, \forall i$. If $\sum \mu_j = 0$, we would have $ \mu_i = 0, \forall i$; however, this cannot be true since we assume that $\mu_i $ are not all zero. Assuming, then, that $\sum \mu_j \neq 0$, we have $\lambda_i = \frac {\mu_i}{\sum \mu_j}$; however, this gives $ \sum \lambda_i = 1 $. Since we assume that $\bm{v}$ is not an affine combination of $\bm{V}$, we arrive at a contradiction and conclude the proof. $\blacksquare$
    
The above theorem shows that the multilateration results in the exact position of the point in the output space when we choose $S$ linearly independent points from the training set and another one (the BAN) that is not an affine combination of the others. Since the number of training points is usually much larger than the dimension of the output, this is usually possible.

            
\subsection{Universal Approximation Property}

We will now verify an important theoretical result for the MLM, its universal approximation capability. The result is divided in two parts: one for the distance estimation error after the linear transformation, and the other for the multilateration estimation error when recovering the output position. This will clarify that the MLM can be used to approximate arbitrary functions.
        

\subsubsection{Upper bound for distance estimation error}

To show that the distance estimation error computed by the MLM is bounded, we will use a result presented in \citep{rbfUA2}, in which the authors show that a Radial Basis Function (RBF) network is a universal approximation. The result is summarized by the following theorem:
		
    	\begin{theorem}[RBF Universal Approximation]
            Let $\kappa: \mathbb{R}^r \rightarrow \mathbb{R}$ be a nonzero integrable function such that $\kappa$ is continuous and radially symmetric with respect to the Euclidean norm. Then the family $S_\kappa$ is dense in the space of continuous $\mathbb{R}$-valued maps defined on any compact subset of $\mathbb{R}^r$ with respect to the norm $||.||_\infty$, where $S_\kappa$ is the family of RBF networks with kernel function $\kappa$ given by $$q(\bm{x}) = \sum_{i = 1}^M w_i\kappa\left(\frac{\bm{x} - \bm{z_i}}{\sigma_i}\right).$$
        \end{theorem}		
		

This theorem shows that an RBF network can approximate a significant set of functions with an arbitrarily small error. For the MLM, we can resort on this result by considering that the desired output of the dataset is the distance to the reference points in the output space. With that modification, we must show that the MLM can be described in the RBF network formalism, which will ensure that the MLM can estimate the distances to the reference points in the output with an arbitrarily small error.

We will first take the centroids of the RBF as the reference points of the MLM. The function $\kappa$ then takes the Euclidean norm, given by $\kappa(\frac{\bm{x} - \bm{z_i}}{\sigma_i}) = ||\frac{\bm{x} - \bm{z_i}}{\sigma_i} ||$. The presented RBF formulation has a parameter $\sigma_i$ that does not appear in the MLM. However, if there is a combination $w_i $, $ \sigma_i$ that satisfies the property, we can calculate $ \bar{w_i} = \frac{w_i}{\sigma_i}$ and get the same result. Finally, both the RBF and the MLM apply a linear regression to compute the output, so we can state that the weights $\bm{w}$ of the RBF are equivalent to the coefficients of matrix $\MatB$ in Eq. \eqref{B}. Thus, we conclude the proof that the error of the MLM-estimated distances in the output space can be arbitrarily small.
        
\subsubsection{Upper bound for the multilateration prediction error}
		
In the previous section, we showed the MLM can provide a good estimate of the distances in the output space. However, the MLM needs an additional step to compute the output: the multilateration. This section shows that the multilateration estimation error is bounded.

In \cite{LLS}, an upper bound is found for the error of multilateration, given by the method detailed in Appendix \ref{ap:LLS}. This work was carried out in the context of localization of mobile autonomous robots and is different in some ways from the MLM. In summary, the objective of that work is to locate a mobile robot based on estimated distances for some fixed points of known locations, called \textit{anchor points}. Both the distance estimates and the anchor point locations themselves may present noise. Thus, in that context, the upper bound for the multilateration error is expressed by the following theorem:
		
\begin{theorem}[Upper bound for the LLS error]
    \label{theorem_bound_lls_error}
	An LLS constructed is described in Eq. \eqref{eq:LLS} in Appendix \ref{ap:LLS} and is expressed
	$$\hat{\bm{A}} \bm{\theta} = \hat{\bm{b}},$$
	where $\hat{\bm{A}} = \bm{A} + \Delta \hat{\bm{A}}$ is a matrix constructed by the anchors' positions, $\bm{A}$ represents the precise position of the anchor nodes, $\Delta \hat{\bm{A}}$ is the anchors' coordinate errors, $\hat{\bm{b}} = \bm{b} + \Delta \hat{\bm{b}}$ is a vector collection of the anchors' positions and the measurement data, $\bm{b}$ denotes the noiseless measurement data, and $\Delta \hat{\bm{b}}$ represents the noise of the measurement data. The ratio between the estimated coordinate $\bm{\hat{y}}$ and the true coordinate $\bm{y}$ satisfies
$$\frac{||\hat{\bm{y}}||}{||\bm{y}||} \leq \psi(1 + \alpha)(1 + \beta),$$
where
    \begin{eqnarray}
		\psi = ||\hat{\bm{A}}^\dag||||\hat{\bm{A}}||, \nonumber \\
        \alpha = \frac{||\Delta \hat{\bm{A}}||}{||\hat{\bm{A}}||}, \nonumber \\
        \beta = \frac{1}{|||\hat{\bm{b}}||_2/||\Delta \hat{\bm{b}}||_2 - 1|}. \nonumber
	\end{eqnarray}
\end{theorem}

An MLM analogy can be made with the presented context by considering that the location of the robot is the desired output and the locations of the anchor points are the locations of the reference points in the output space. Distance estimates from the robot to the anchor points are given by the output of the MLM before the multilateration step. 

Theorem \ref{theorem_bound_lls_error} presents an upper bound for the multilateration error. However, the characteristics of the MLM allows us to make the bound tighter. First, we consider that the location of the reference points is accurate, which means that $\Delta\bm{A} = \bm{0}$; thus, $\alpha = 0$. In addition, we saw in the previous section that the MLM distance estimate errors can be
arbitrarily small. This means $ \Delta\bm{b} \rightarrow \bm{0}$, which implies that $\beta \rightarrow 0 $. Thus, we can present the following corollary:

\begin{corollary}
    \label{corollary_bound_lls_error}
    The error of the MLM multilateration step is bounded by $$\frac{||\hat{\bm{y}}||}{||\bm{y}||} \leq \psi(1 + \beta) = \mathcal{U},$$
where
    \begin{eqnarray}
		\psi = ||\hat{\bm{A}}^\dag||||\hat{\bm{A}}||, \nonumber \\
        \beta = \frac{1}{|||\hat{\bm{b}}||_2/||\Delta \hat{\bm{b}}||_2 - 1|}. \nonumber
	\end{eqnarray}
	
In addition, since we have $ \beta \rightarrow 0 $, we have $ \mathcal{U} \rightarrow \psi $.
\end{corollary}

The result of this corollary shows that the ratio between the returned and the desired output is bounded. We will develop this relation to show that the distance between them is also bounded:
		
    \begin{align}
        \nonumber
		d(\bm{\hat y}, \bm{y})^2 = & (\bm{\hat y} - \bm{y})^T(\bm{\hat y} - \bm{y})  \\
        \nonumber
        d(\bm{\hat y}, \bm{y})^2 = & ||\bm{\hat y}||^2 + ||\bm{y}||^2 -2\bm{y}^T\bm{\hat y} \\
        \nonumber
        d(\bm{\hat y}, \bm{y})^2 = & ||\bm{\hat y}||^2 + ||\bm{y}||^2 -2||\bm{y}||||\bm{\hat y}||\cos \alpha \\
        \nonumber
        \frac{d(\bm{\hat y}, \bm{y})^2}{||\bm{y}||^2} = & \frac{||\bm{\hat y}||^2}{||\bm{y}||^2} + \frac{||\bm{y}||^2}{||\bm{y}||^2} -2\cos \alpha\frac{||\bm{y}||||\bm{\hat y}||}{||\bm{y}||^2} \\ \nonumber
        \frac{d(\bm{\hat y}, \bm{y})^2}{||\bm{y}||^2} = & \left(\frac{||\bm{\hat y}||}{||\bm{y}||}\right)^2 + 1 -2\cos \alpha\frac{||\bm{\hat y}||}{||\bm{y}||} \\
        \label{eq:mult_dist}
        \frac{d(\bm{\hat y}, \bm{y})^2}{||\bm{y}||^2} \leq & (\mathcal{U})^2 + 1 -2\cos \alpha(\mathcal{U}).
	\end{align}		
		
We can therefore conclude that if the norm $||\bm{y}||$ of the target output is bounded, the distance $d(\bm{\hat y}, \bm{y})$ between the desired output and the output estimated by the multilateration is also bounded.
		
\subsection{Discussion}

Corollary \ref{corollary_bound_lls_error} indicates the upper bound of the multilateration error depends on matrix $\bm{A}$, which itself is associated with the reference points used to compute the distances. This observation indicates we can make the bound tighter for certain choices of reference points, thereby reducing the output estimation error limit. This idea was previously demonstrated empirically \citep{RSesann2018,florencio2018, Maia2018}. In the present work, we have now theoretically motivated a non-random selection for the reference points. We assess that motivation by performing comprehensive computational experiments detailed in the next sections, with a focus on clustering-based approaches.

\section{Clustering-based Selection of Reference Points}
\label{prop}

In this section, we evaluate four clustering-based methods in the reference point selection problem. The methods include two nondeterministic and two deterministic ones. A general algorithm for the selection of clustering-based reference points is depicted in Algorithm \ref{alg:RS}. All the methods are based on a common strategy, where the selection of reference points is performed only in the input space. The corresponding points (indices) are simply selected as output references. Therefore, in the following, we consider only the input space when describing the proposed methods.

\begin{algorithm}[t]
\caption{Clustering-based selection of reference points}
\label{alg:RS}
\begin{algorithmic}[1]
\REQUIRE input points $\SetX$, output points $\SetY$, and number of reference points $K$.
\ENSURE reference points $\SetR$ and $\SetT$.
	\STATE Cluster $\SetX$ to $K$ clusters.
	\STATE Select cluster prototype from each cluster
	\STATE Select $\SetR$ according to the cluster prototypes from $\SetX$
	\STATE Select $\SetT$ corresponding to indices of $\SetR$ from $\SetY$.
 \end{algorithmic}
\end{algorithm}

\subsection{Methods}

\begin{table}[!t]
\begin{adjustbox}{max width=\textwidth}
\centering
\begin{tabular}{lllll}
\hline
\textbf{Method} & \textbf{Based on}  & \textbf{Deterministic}  & \textbf{Type} & \textbf{Complexity} \\ \hline
{RS-K-means++} & K-means++ initialization & No & Partitional &  $\mathcal{O}(N)$\\
{RS-K-medoids++} & K-means++ initialization and & No & Partitional &  $\mathcal{O}(N)$\\
 & K-medoids clustering &  \\
{RS-UPGMA} & Aggloremerative clustering & Yes & Hierarchical & $\mathcal{O}(N^2)$\\
{RS-maximin} & Maximin clustering initialization & Yes & Partitional & $\mathcal{O}(N)$\\
\hline 
\end{tabular}
\end{adjustbox}
\caption{Summary of the evaluated reference point selection approaches.}
\label{tab:methods}
\end{table}

The K-means++ initialization method \citep{arthur2007k} is one of the most popular methods of K-means initialization. The first method we evaluate is the use the K-means++ initialization with the Euclidean distance for the selection of reference points. See \cite{hamalainen2017comparison} for a description of the algorithm. We will refer to this approach as reference point selection with K-means++ (RS-K-means++).

The second evaluated approach begins by running the K-means++ initialization with the Euclidean distance and then refines the initial prototypes with Lloyd’s algorithm \citep{lloyd1982least} until convergence. Finally, the closest observation to each final prototype (medoid) is picked as the reference point. These closest points then establish the set of selected reference points. This method is referred to as RS-K-medoids++. Both RS-K-medoids++ and RS-K-means++ are nondeterministic methods, because of the random sampling of the initial prototypes based on the Euclidean distance-constructed probability distribution (see \citealt{hamalainen2017comparison} and articles therein).

The unweighted pair group method with arithmetic mean (UPGMA; \citealt{sokal1958statistical}) is an agglomerative clustering algorithm that starts clustering from the initial state, where each point forms one cluster. Then, in each step, the two clusters that have the smallest average distance between the cluster members are joined together. The third evaluated method utilizes UPGMA on the data, and then computes the mean prototypes for each cluster; finally, it again selects the closest point to the prototype as a reference point. Similar to RS-K-medoids++, those closest points construct the set of selected reference points. We refer to this method as RS-UPGMA.

The fourth evaluated method is based on a maximin clustering initialization algorithm \citep{gonzalez1985clustering}. The original method starts with a random initial point and then picks each new point, similar to the K-means++ method. However, unlike K-means++, a point that has the farthest distance to the closest already selected point is chosen as a new point. Our modification of the maximin first selects the closest point to the data mean as the first point, conceiving the whole algorithm as completely deterministic. This approach is referred to as RS-maximin. We emphasize that the latter two approaches, RS-UPGMA and RS-maximin, are deterministic.

One of the justifications for selecting this specific set of clustering methods is the highly different amounts of separation between the selected reference points (see Figure \ref{fig:S1RSdists} in Appendix \ref{app:figures}). Random selection has the smallest amount of separation among the reference points, and the RS-maximin method has the largest; RS-K-means++, RS-K-medoids++, and RS-UPGMA interpolate between these two extremes. There are plenty of clustering methods available; the methods evaluated here are straightforward and easy to implement. Moreover, the MLM has only one hyperparameter, the number of reference points $K$ to be selected, which the methods keep unchanged. 

A summary of the evaluated approaches is shown in Table \ref{tab:methods}, where the time complexities are also presented with respect to the number of training observations $N$. RS-K-means++, RS-K-medoids++, and RS-maximin have linear time complexity. The UPGMA has quadratic complexity \citep{gronau2007optimal}; therefore, the complexity of RS-UPGMA is also quadratic, since the post-processing after the UPGMA clustering step has linear time complexity. Since the MLM training phase has a time complexity of $\mathcal{O}(K^2N)$ \citep{de2015minimal}, a reference point selection method with a linear computational cost (with respect to $N$ ) and the ability to build an accurate model with a small $K$ is highly desirable.

\subsection{Motivation for Reference Point Selection}\label{subsec:RefPointMot}

%
%

\begin{figure}[!t]
     \centering
     \begin{subfigure}[b]{0.49\textwidth}
         \centering
         \includegraphics[width=\textwidth]{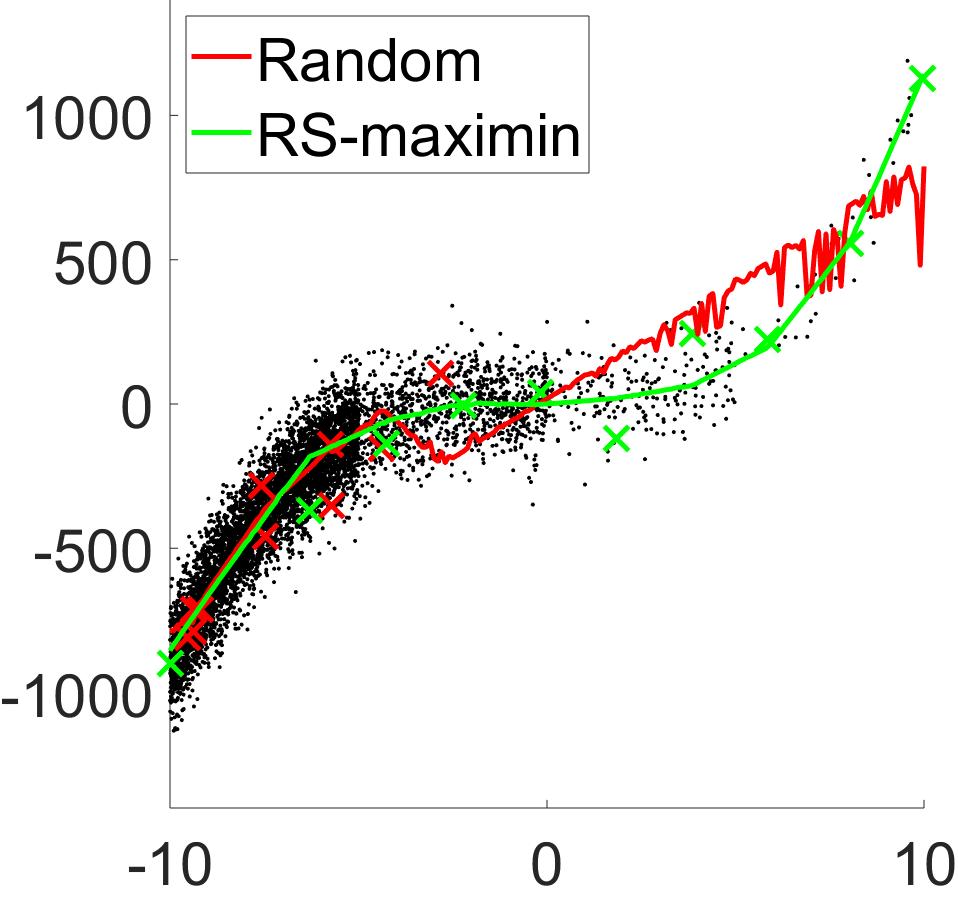}
         \label{fig:randSel}
     \end{subfigure}
     \hfill
     \begin{subfigure}[b]{0.49\textwidth}
         \centering
         \includegraphics[width=\textwidth]{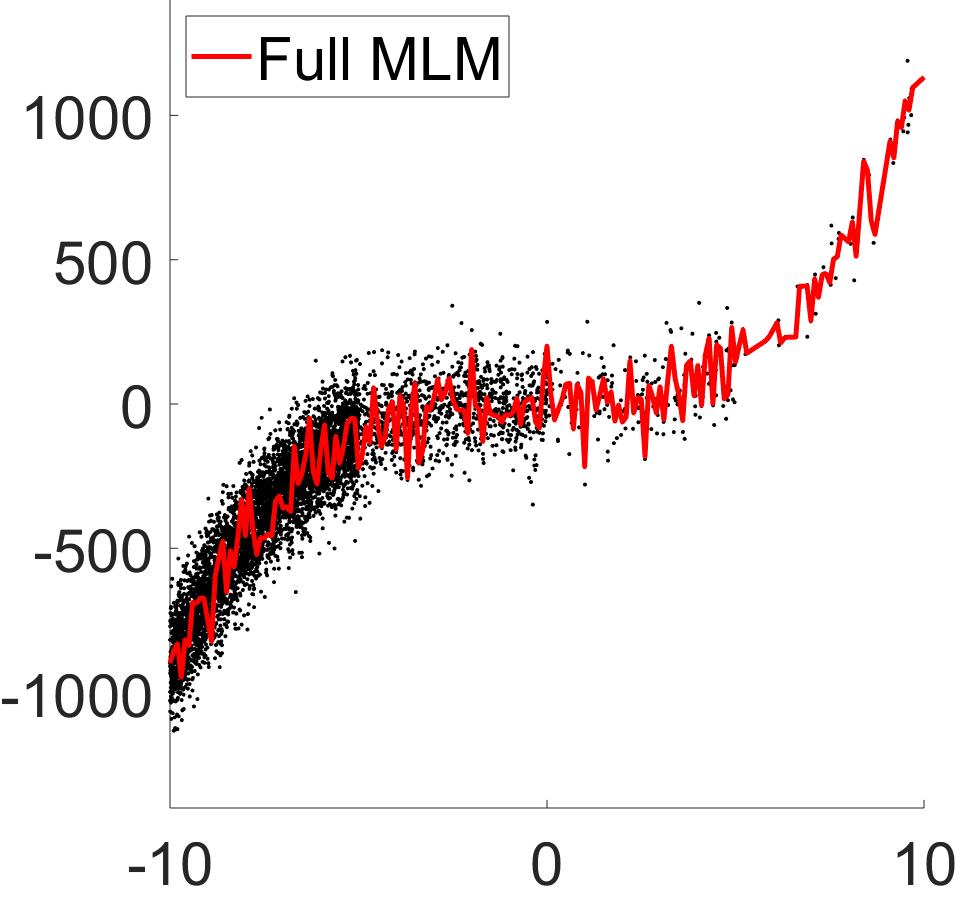}
         \label{fig:clusterSel}
     \end{subfigure}
     \caption{Illustration of effects for different reference selection strategies. Selected reference points are marked with crosses with corresponding color of the fitted curve.}
     \label{fig:toydataset}
\end{figure}

To illustrate reference point selection effects in terms of MLM's accuracy, we generated a nonlinear synthetic dataset (6240 observations, 1 input variable, 1 output variable) with varying density. Input values are drawn from four highly different density intervals. Corresponding output values are given by a cubic function with gaussian noise. MLM was trained with the Random and RS-maximin methods when $K = 10$. In addition, we trained the Full MLM variant. Figure \ref{fig:toydataset} illustrates that the Random method selects reference points from high-density regions, which causes MLM to have a very low accuracy in low-density regions. RS-maximin selects reference points also from the low-density regions, which clearly improves the accuracy. Selecting reference points from near the data cloud boundaries improves the extrapolation capability of the MLM regression model as also illustrated in (\citealt[pp. 35]{hamalainen2018improvements}). A straightforward approach to cover also low-density regions is to include all data points as a reference points, however, this can lead to overfitting and very large MLM models. On the other hand, overfitting seems to be rarely a problem for multidimensional input spaces based on the experimental results of this paper and works of \citep{florencio2020new, HamKar2020, KarNeCo2019, pihlajamaki2020monte}. Especially in classification problems, small noise on the class boundaries as characterized in Figure \ref{fig:toydataset} (right) might not affect the classification accuracy.

\section{Experiments and Results}
\label{exp}

In this section, we show empirical evidence with an extensive set of datasets how the MLM's generalization accuracy can be improved in regression problems with the clustering-based selection of the reference points. We used the Random selection as a baseline for the clustering-based methods.

\subsection{Experimental Setup}

\begin{table}[!t]
\centering
\begin{tabular}{lrr}
\hline
\textbf{Dataset} & \textbf{\# Observations}  & \textbf{\# Features} \\ \hline
Auto Price (AP) & 159 & 15 \\
Servo (SRV) & 167 & 4 \\
Breast Cancer (BC) & 194 & 32 \\
Computer Hardware (CHA) & 209 & 6 \\
Boston Housing (BH) & 506 & 13 \\
Forest Fires (FF) & 517 & 12 \\
Stocks (STC) & 950 & 9 \\
S1 (S1) & 1000 & 2 \\
Bank (BNK) & 4499 & 8 \\
Ailerons (ALR) & 7129 & 5 \\
Computer Activity (CA) & 8192 & 12 \\
Elevators (ELV) & 9517 & 6 \\
Combined Cycle Power Plant (CCP) & 9568 & 4 \\
California Housing (CH) & 20640 & 8 \\
Census (CNS) & 22784 & 8 \\
\hline 
\end{tabular}
\caption{Characteristics of the datasets used in the experiments.}
\label{tab:datasets}
\end{table}

We selected 13 real datasets and two synthetic datasets (S1, BNK) to evaluate the reference point selection methods. The selected datasets are summarized in Table \ref{tab:datasets}. All datasets had one-dimensional output values. The S1 dataset was modified for a regression task. We randomly selected 1000 observations from the original S1 data, scaled their values to the range $[0,1]$ and then computed the output values $f(x_1,x_2)$ with the function $\sin(2\pi x_1) + \sin(2\pi x_2)$. The original S1 dataset is available in (\url{http://cs.uef.fi/sipu/datasets/}). The remaining datasets are available at (\url{http://www.dcc.fc.up.pt/~ltorgo/Regression/DataSets.html}) and at (\url{http://archive.ics.uci.edu/ml/index.php}).

For a more rigorous comparison, we performed model selection and assessment as follows: we divided the original datasets into train-validation-test sets and performed cross-validation (see, e.g., \citealt{friedman2001elements}, Chapter 7). More precisely, we used the 3-DOB-SCV \citep{Mor2012} approach to divide each dataset into a training and a test set. Therefore, the test set was forced to approximate the same distribution as the training set, thus making the comparison more reliable in the event concept drift is not considered. Because we focused only on regression tasks, we used DOB-SCV as a one-class case \citep{hamalainen2018improvements,HamKar2016}.  Moreover, we archived three training sets and three test sets for each dataset, respectively, with sizes of $2/3$ and $1/3$ of the number of observations. In training, we used the 10-DOB-SCV approach to select the optimal number of reference points. Hence, $18/30$ of the number of observations was used to train the model and $2/30$ of the number of observations was used to compute the validation error. Therefore, we have a two-level division of the datasets.

We evaluated the quality of the models using the root mean squared error (RMSE). In addition to the validation error, a test error was also computed for all 10-DOB-SCV training sets, resulting in 10 test RMSEs for each training set and 30 test RMSEs for the overall dataset. For more interpretable results, we expressed the number of the selected reference points in a relative manner:
\begin{equation}\label{Krel}
K_{rel} = 100 \frac{K}{N},
\end{equation}
where $N$ is the number of observations in the training data. In training, the number of reference points $K_{rel}$ varied in the range of $[5,100]$, with a step size of $5$. We used  the LLS method for the output prediction (Algorithm \ref{alg:OutPredLLS}).  To solve the linear system of equations in the MLM implementation, we utilized MATLAB’s {\sl{mldivide}}-function. We scaled all training observations to the range $[0,1]$. All the experiments were conducted in a MATLAB environment.

\subsection{Results for Optimal $K$}

\begin{table}[!t]
\huge
\centering
\bgroup
\def\arraystretch{1.5}%
\begin{adjustbox}{max width=\textwidth}
\begin{tabular}{llllllllllllllllllll}
\hline
\multirow{ 2}{*}{} & \multicolumn{2}{l}{\textbf{Random}} & \multicolumn{2}{l}{\textbf{RS-K-means++}} & \multicolumn{2}{l}{\textbf{RS-K-medoids++}} & \multicolumn{2}{l}{\textbf{RS-UPGMA}} & \multicolumn{2}{l}{\textbf{RS-maximin}} \\
\textbf{Dataset}   & \textbf{RMSE}  & $\mathbf{K_{rel}}$ & \textbf{RMSE} & $K_{rel}$  & \textbf{RMSE}  & $\mathbf{K_{rel}}$ & \textbf{RMSE}  & $\mathbf{K_{rel}}$ & \textbf{RMSE}  & $\mathbf{K_{rel}}$\\ \hline
AP & $0.0640(85)$ & $75,90,80$ & $0.0600(77)$ & $65,80,95$ & $0.0597(79)$ & $\bf{45,100,55}$ & $\bf{0.0583}(68)$ & $90,95,100$ & $\bf{0.0583}(68)$ & $90,100,95$\\ 
SRV & $0.0920(77)$ & $\bf{90,100,90}$ & $0.0910(76)$ & $90,100,95$ & $0.0908(76)$ & $95,100,95$ & $\bf{0.0899}$$(75)$ & $100,100,95$ & $0.0913$$\bf{(74)}$ & $95,100,100$\\ 
BC & $0.2678(78)$ & $10,20,30$ & $0.2674(85)$ & $5,25,15$ & $0.2664(76)$ & $\bf{10,15,10}$ & $0.2680(77)$ & $10,15,15$ & $\bf{0.2647}(63)$ & $15,20,10$\\ 
CHA$^{**}$ & $0.0483(93)$ & $80,15,80$ & $0.0461(98)$ & $\bf{50,15,25}$ & $0.0431(66)$$^{\dagger}$ & $25,65,55$ & $\bf{0.0403}(52)$$^{*\dagger}$ & $20,95,20$ & $0.0421(68)$ & $65,55,15$\\ 
BH & $0.0728(81)$ & $75,100,100$ & $\bf{0.0717}(72)$ & $95,100,100$ & $0.0725$$\bf{(75)}$ & $95,85,100$ & $\bf{0.0717}(75)$ & $\bf{85,85,75}$ & $0.0718$$\bf{(75)}$ & $85,95,100$\\ 
FF & $\bf{0.0557}$$(74)$ & $40,10,10$ & $0.0564(82)$ & $5,45,10$ & $0.0559(75)$ & $30,10,55$ & $0.0566$$\bf{(71)}$ & $\bf{5,5,15}$ & $0.0566(76)$ & $\bf{5,15,5}$\\ 
STC & $\bf{0.0227}(73)$ & $100,100,100$ & $0.0228(77)$ & $\bf{95,100,95}$ & $0.0228(74)$ & $100,95,100$ & $\bf{0.0227}(75)$ & $\bf{100,100,90}$ & $0.0228(78)$ & $\bf{95,100,95}$\\ 
S1 & $\bf{0.0051}(76)$ & $100,100,100$ & $0.0053(76)$ & $75,100,95$ & $0.0052(78)$ & $80,90,70$ & $\bf{0.0051}$$(75)$ & $\bf{80,90,65}$ & $0.0052$$\bf{(74)}$ & $85,100,95$\\ 
BNK$^{**}$ & $0.0514(95)$ & $90,95,90$ & $0.0515(93)$ & $85,100,70$ & $0.0509(81)$ & $100,55,60$ & $0.0490(59)$$^{*\dagger}$ & $95,5,10$ & $\bf{0.0481}(51)$$^{*\dagger}$ & $\bf{5,10,10}$\\ 
ALR & $\bf{0.0417}(66)$ & $10,10,10$ & $0.0418(71)$ & $5,10,15$ & $0.0418(69)$ & $10,5,15$ & $0.0420(87)$ & $5,10,20$ & $0.0420(85)$ & $\bf{5,10,5}$\\ 
CA & $\bf{0.0288}$$(75)$ & $90,100,100$ & $\bf{0.0288}$$(75)$ & $90,75,75$ & $\bf{0.0288}(72)$ & $60,85,75$ & $0.0289(79)$ & $\bf{70,65,60}$ & $\bf{0.0288}$$(77)$ & $80,95,70$\\ 
ELV & $0.0554(76)$ & $\bf{5,5,5}$ & $\bf{0.0553}(74)$ & $\bf{5,5,5}$ & $0.0554(75)$ & $\bf{5,5,5}$ & $\bf{0.0553}(75)$ & $\bf{5,5,5}$ & $\bf{0.0553}$$(76)$ & $5,5,10$\\ 
CCP & $0.0478$$\bf{(73)}$ & $85,100,100$ & $\bf{0.0476}(73)$ & $90,70,95$ & $0.0480(77)$ & $70,85,95$ & $0.0482(81)$ & $\bf{55,80,95}$ & $0.0480(74)$ & $75,80,95$\\ 
CH & $0.1137(77)$ & $\bf{70,85,70}$ & $\bf{0.1134}$$(75)$ & $80,80,95$ & $0.1135(76)$ & $80,80,95$ & $0.1135(75)$ & $100,95,100$ & $0.1136$$(\bf{74})$ & $100,100,90$\\ 
CNS & $0.0605(83)$ & $20,35,30$ & $0.0605(80)$ & $15,20,20$ & $0.0603(75)$ & $15,25,25$ & $\bf{0.0599}(64)$ & $15,25,15$ & $0.0602(76)$ & $\bf{10,30,10}$\\ 
\hline
\multirow{ 1}{*}{Rank $/$ $K_{rel}^{avg}$} & \multicolumn{2}{l}{5(54) $/$ 64.44 } & \multicolumn{2}{l}{4(48) $/$ 59.56} &
\multicolumn{2}{l}{3(44) $/$ 58.44} & \multicolumn{2}{l}{2(40) $/$ \textbf{55.22}} & \multicolumn{2}{l}{\textbf{1(39)} $/$ 56.33} \\ \hline 
\end{tabular}
\end{adjustbox}
\egroup
\caption{RMSE for the optimal $K$} \label{ValidationErr}
\end{table}

Table \ref{ValidationErr} shows the median test RMSE and the best number of reference points. The optimal number of reference points was selected based on the smallest mean validation RMSE. The symbol $**$ indicates a statistically significant difference between test RMSEs, based on a Kruskal-Wallis H test with a significance level of $0.05$. The symbols $*$, $\dagger$, $\ddagger$, $\mathsection$, and $\|$ denote that a method has a statistically significantly smaller RMSE in pairwise comparison to Random, RS-K-means++, RS-K-medoids++, RS-UPGMA, and RS-maximin, respectively. In the pairwise comparisons, the significance level was also set to $0.05$. The Kruskal-Wallis H test assumes equal variances for groups; therefore, we tested equality of variances with a Brown-Forsythe test. Based on that test, variances related to optimal $K$ results are equal for all datasets. The best median test RMSE and the set of the smallest number of reference points (with respect to the mean value) are in boldface for each dataset. Note that in Table \ref{ValidationErr}, there are 3 optimal $K$ values for each method, since we used the 3-DOV-SCV approach in the experiments. Rounded Kruskal-Wallis scores are shown inside the brackets and the best scores are in boldface. Dataset-wise ranking of the methods is calculated from the raw Kruskal-Wallis scores. Based on these rankings, the final ranking of the methods is shown at the bottom of Table \ref{ValidationErr}. In addition, the average $K_{rel}$ is also shown at the bottom of Table \ref{ValidationErr} for each method.

Based on Table \ref{ValidationErr}, RS-UPGMA and RS-maximin perform equally well in the final ranking, while RS-K-medoids++ and RS-K-means++ perform similarly. In terms of the final ranking and the model size ($K_{rel}$), Random has the worst performance and the deterministic methods RS-UPGMA and RS-maximin have the best performance. In general, clustering-based methods give sparser models that reduce the computational cost and the space complexity. In addition, the clustering-based models have better generalization ability. Based on the Kruskal-Wallis test, there are statistically significant differences between the methods for the CHA and BNK datasets in favor of the deterministic methods. For the BNK dataset, RS-maximin builds the MLM model with only $K_{rel} = \{5,10,10\}$, while Random must select almost the entire dataset as reference points ($K_{rel} = \{90,95,90\}$) and still has a clearly larger RMSE error. Reducing $K_{rel}$ from 90 to 10 reduces space requirements for the distance regression model coefficient matrix by $98.77\%$. For large datasets where $N >>P$ and $N >> L$, this coefficient matrix size determines the space complexity of the full MLM model ($K_{rel} = 100$).



The best $K$ selection based on the smallest mean validation RMSE is dubious for some of the datasets, since the complexity of the model is not taken into account. For example, for a large dataset, if increasing $K_{rel}$ from 50 to 100 leads to only marginal improvement in the mean validation RMSE, then the model with higher $K$ and smaller mean validation RMSE is selected. For example, for the S1 dataset, RS-maximin already achieves the fulll MLM error level when $K_{rel} = 20-40$ (see Tables \ref{ValidationErrK20} and \ref{ValidationErrK40}). For future work, there is still room for improvement in this respect.

\begin{table}[!t]
\centering
\bgroup
\def\arraystretch{1.0}%
\begin{adjustbox}{max width=.8\textwidth}
\begin{tabular}{lllllllllllll}
\hline
\multirow{ 1}{*}{Dataset} & \multicolumn{1}{l}{\textbf{Random}} & \multicolumn{1}{l}{\textbf{RS-K-means++}} & \multicolumn{1}{l}{\textbf{RS-K-medoids++}} & \multicolumn{1}{l}{\textbf{RS-UPGMA}} & \multicolumn{1}{l}{\textbf{RS-maximin}} \\ \hline
AP$^{**}$ & $0.1083(89)$ & $0.1082(83)$ & $0.1052(86)$ & $0.0954(66)$ & $\bf{0.0829}(54)$$^{*\ddagger}$\\ 
SRV$^{**}$  & $\bf{0.1921}(57)$$^{\|}$ & $0.2024(70)$ & $0.2088(86)$ & $0.2011(71)$ & $0.2132(94)$\\ 
BC & $0.2672$$\bf{(68)}$ & $0.2684(80)$ & $\bf{0.2670}$$(71)$ & $0.2707(86)$ & $0.2671(72)$\\ 
CHA & $0.0697(82)$ & $\bf{0.0593}(57)$ & $0.0659(80)$ & $0.0682(78)$ & $0.0608(81)$\\ 
BH & $0.1171(70)$ & $0.1194(80)$ & $0.1141(84)$ & $0.1141(80)$ & $\bf{0.1099}(63)$\\ 
FF & $0.0572(81)$ & $\bf{0.0565}$$(76)$ & $0.0568(81)$ & $0.0566$$\bf{(67)}$ & $0.0566(73)$\\ 
STC$^{**}$ & $0.0521(121)$ & $0.0478(87)$$^{*}$ & $0.0457(45)$$^{*\dagger\|}$ & $\bf{0.0449}(41)$$^{*\dagger\|}$ & $0.0477(84)$$^{*}$\\ 
S1$^{**}$ & $0.0366(128)$ & $0.0285(91)$$^{*}$ & $0.0270(78)$$^{*}$ & $0.0241(56)$$^{*\dagger}$ & $\bf{0.0199}(25)$$^{*\dagger\ddagger\mathsection}$\\ 
BNK$^{**}$ & $0.0645(103)$ & $0.0584(87)$ & $0.0670(117)$ & $0.0499(42)$$^{*\dagger\ddagger}$ & $\bf{0.0491}(30)$$^{*\dagger\ddagger}$\\ 
ALR & $\bf{0.0417}(70)$ & $0.0418(68)$ & $0.0419(77)$ & $0.0420(83)$ & $0.0420(79)$\\ 
CA$^{**}$ & $0.0341(120)$ & $0.0320(70)$$^{*}$ & $\bf{0.0314}$$(54)$$^{*\mathsection}$ & $0.0324(85)$$^{*}$ & $\bf{0.0314}(48)$$^{*\mathsection}$\\ 
ELV & $0.0554(77)$ & $\bf{0.0553}$$(75)$ & $0.0554(76)$ & $\bf{0.0553}$$(76)$ & $\bf{0.0553}(73)$\\ 
CCP & $0.0528(88)$ & $0.0526(79)$ & $0.0526(73)$ & $0.0525(70)$ & $\bf{0.0522}(67)$\\ 
CH$^{**}$  & $\bf{0.1201}(44)$$^{\ddagger\mathsection\|}$ & $0.1208(58)$$^{\mathsection\|}$ & $0.1219(75)$ & $0.1230(101)$ & $0.1232(99)$\\ 
CNS & $0.0622(91)$ & $0.0613(77)$ & $0.0614(79)$ & $\bf{0.0607}$$(68)$ & $\bf{0.0607}(62)$\\ 
\hline 
\multirow{ 1}{*}{Rank} & 
\multicolumn{1}{l}{5(55)} & \multicolumn{1}{l}{2(42)} &
\multicolumn{1}{l}{4(52)} & \multicolumn{1}{l}{3(43)} & \multicolumn{1}{l}{\bf{1(33)}} \\ \hline
\end{tabular}
\end{adjustbox}
\egroup
\caption{RMSE for $K_{rel} = 5$} \label{ValidationErrK5}
\end{table}

\begin{table}[!t]
\centering
\bgroup
\def\arraystretch{1}%
\begin{adjustbox}{max width=.8\textwidth}
\begin{tabular}{lllllllllllll}
\hline
\multirow{ 1}{*}{Dataset} & \multicolumn{1}{l}{\textbf{Random}} & \multicolumn{1}{l}{\textbf{RS-K-means++}} & \multicolumn{1}{l}{\textbf{RS-K-medoids++}} & \multicolumn{1}{l}{\textbf{RS-UPGMA}} & \multicolumn{1}{l}{\textbf{RS-maximin}} \\ \hline
AP & $0.0930(83)$ & $0.0916(85)$ & $0.0856(78)$ & $0.0838(70)$ & $\bf{0.0762}(62)$\\ 
SRV$^{**}$ & $0.1479(64)$ & $0.1500(60)$ & $0.1651(91)$ & $0.1693(108)$ & $\bf{0.1468}(54)$\\ 
BC & $0.2674(77)$ & $0.2667(72)$ & $0.2668(77)$ & $0.2692(85)$ & $\bf{0.2655}(67)$\\ 
CHA$^{**}$ & $0.0613(105)$ & $0.0542(92)$ & $0.0496(81)$ & $\bf{0.0428}(43)$$^{*\dagger\ddagger}$ & $0.0448(57)$$^{*\dagger}$\\ 
BH & $\bf{0.0994}(70)$ & $0.1003(76)$ & $0.1017(75)$ & $0.1018(81)$ & $0.1011(75)$\\ 
FF & $0.0568(78)$ & $0.0568(79)$ & $0.0572(80)$ & $0.0567(70)$ & $\bf{0.0565}(72)$\\ 
STC$^{**}$ & $0.0395(122)$ & $0.0375(102)$ & $0.0359(59)$$^{*\dagger}$ & $\bf{0.0353}(45)$$^{*\dagger}$ & $0.0359(50)$$^{*\dagger}$\\ 
S1$^{**}$ & $0.0188(123)$ & $0.0140(92)$$^{*}$ & $0.0135(81)$$^{*}$ & $0.0109(60)$$^{*\dagger}$ & $\bf{0.0078}(23)$$^{*\dagger\ddagger\mathsection}$\\ 
BNK$^{**}$ & $0.0589(103)$ & $0.0570(93)$ & $0.0588(97)$ & $0.0487(47)$$^{*\dagger}$ & $\bf{0.0481}(37)$$^{*\dagger}$\\ 
ALR & $\bf{0.0417}(65)$ & $0.0418(72)$ & $0.0418(63)$ & $0.0420(86)$ & $0.0422(92)$\\ 
CA$^{**}$ & $0.0316(121)$ & $0.0302(73)$$^{*}$ & $\bf{0.0299}(61)$$^{*}$ & $0.0305(71)$$^{*}$ & $0.0301(52)$$^{*}$\\ 
ELV & $0.0557(85)$ & $0.0555(83)$ & $0.0555(76)$ & $0.0555(76)$ & $\bf{0.0554}(66)$\\ 
CCP & $0.0516(88)$ & $0.0516(78)$ & $0.0515(72)$ & $0.0515(71)$ & $\bf{0.0511}(68)$\\ 
CH$^{**}$ & $\bf{0.1177}(44)$$^{\mathsection\|}$ & $0.1185(59)$$^{\mathsection\|}$ & $0.1193(74)$ & $0.1212(100)$ & $0.1212(101)$\\ 
CNS & $0.0612(89)$ & $0.0605(78)$ & $0.0605(80)$ & $\bf{0.0601}$$(67)$ & $0.0602$$(\bf{64})$\\ 
\hline 
\multirow{ 1}{*}{Rank} & \multicolumn{1}{l}{5(58)} & \multicolumn{1}{l}{4(52)} &
\multicolumn{1}{l}{3(46)} & \multicolumn{1}{l}{2(41)} & \multicolumn{1}{l}{\bf{1(28)}} \\ \hline
\end{tabular}
\end{adjustbox}
\egroup
\caption{RMSE for $K_{rel} = 10$} \label{ValidationErrK10}
\end{table}

\subsection{Results for Fixed $K$}
\label{Sect:fixedK}


Tables \ref{ValidationErrK5}--\ref{ValidationErrK40} show the test RMSEs. They are similar to Table \ref{ValidationErr}, but with a fixed number of reference points. Variances for the error distributions are not equal for SRV ($K_{rel} = 5,10,20$), CHA ($K_{rel} = 10,20,40$), BH ($K_{rel} = 10,20$), FF ($K_{rel} = 10$), STC ($K_{rel} = 5,10,20,40$), S1 ($K_{rel} = 5,10,20$), CA ($K_{rel} = 5,20,40$), and ELV ($K_{rel} = 10$), based on the Brown-Forsythe test of group variances. Therefore, the results given by Kruskal-Wallis are questionable for these cases. However, the ordering of the methods can still be compared.

As expected, based on the final ranking, all the proposed methods have better RMSE than Random when the number of reference points is small to moderate ($K_{rel} = 5,10,20,40$, Tables \ref{ValidationErrK5}--\ref{ValidationErrK40}). RS-K-means++ have better RMSEs compared to RS-K-medoids++ for $K_{rel} = 5$. Thus, refinement of the reference points with K-means does not seem to be beneficial for the small $K_{rel}$. In contrast to $K_{rel}= 10, 20$, accuracy is improved with K-means refinement. In general, the RS-maximin method obtained the best RMSE in the comparison. RS-UPGMA have results that are quite similar to those of RS-maximin for $K_{rel} = 20, 40$. Therefore, running the whole clustering (not only the initialization step) seems to work better for higher $K$ values. For $K_{rel} = 20$, RS-UPGMA is the best approach based on the final ranking.

A drawback of RS-K-medoids++, RS-UPGMA, and RS-maximin is that if the data contains anomalies, they are prone to select them as reference points. This is probably the reason why Random gets smaller RMSE than RS-UPGMA and RS-maximin for the CH dataset with small to moderate $K_{rel}$, since that dataset is known to contain some large anomalies. Therefore, we combined a simple anomaly detection method (k-nearest neighbors) with RS-UPGMA and tested it with the CH dataset. It was observed that anomaly detection improved the test error for RS-UPGMA ($K_{rel}=5,10,20,40$). Similar observations can also be drawn from the results for the S1 dataset. S1 is the cleanest dataset in our experiments: all input points are mapped to output points with sine-based function evaluations without any distortions. Based on Tables \ref{ValidationErrK5}--\ref{ValidationErrK40}, RS-UPGMA and RS-maximin have the largest error differences compared to Random for the S1 dataset than any other dataset. Therefore, a robust variant of the MLM combined with RS-UPGMA or RS-maximin should be considered for regression tasks with anomalies.

\subsection{Case S1: Comparison of Methods}

\begin{table}[!t]
\centering
\bgroup
\def\arraystretch{1}%
\begin{adjustbox}{max width=.8\textwidth}
\begin{tabular}{lllllllllllll}
\hline
\multirow{ 1}{*}{Dataset} & \multicolumn{1}{l}{\textbf{Random}} & \multicolumn{1}{l}{\textbf{RS-K-means++}} & \multicolumn{1}{l}{\textbf{RS-K-medoids++}} & \multicolumn{1}{l}{\textbf{RS-UPGMA}} & \multicolumn{1}{l}{\textbf{RS-maximin}} \\ \hline
AP & $0.0858(87)$ & $0.083(82)$ & $0.0794(72)$ & $0.0775(68)$ & $\bf{0.0738}(68)$\\ 
SRV & $0.1226(73)$ & $0.1244(\bf{71})$ & $\bf{0.1213}(71)$ & $0.1239(82)$ & $0.1225(80)$\\ 
BC & $0.2671(80)$ & $\bf{0.2651}(70)$ & $0.2659(80)$ & $0.2655(73)$ & $0.2655(74)$\\ 
CHA$^{**}$ & $0.0595(111)$ & $0.0475(95)$ & $0.0443(72)$$^{*}$ & $\bf{0.0403}(45)$$^{*\dagger}$ & $0.0430(55)$$^{*\dagger}$\\ 
BH & $0.0913(82)$ & $0.0875(79)$ & $0.0890(81)$ & $\bf{0.0836}(58)$ & $0.0857(78)$\\ 
FF & $0.0565(80)$ & $0.0564(76)$ & $0.0568(82)$ & $\bf{0.0560}(70)$ & $0.0565$$\bf{(70)}$\\ 
STC$^{**}$ & $0.0324(129)$ & $0.0304(95)$$^{*}$ & $0.0296(83)$$^{*}$ & $0.0283(42)$$^{*\dagger\ddagger}$ & $\bf{0.0277}(29)$$^{*\dagger\ddagger}$\\ 
S1$^{**}$ & $0.0113(128)$ & $0.0082(88)$$^{*}$ & $0.0083(83)$$^{*}$ & $0.0069(49)$$^{*\dagger\ddagger}$ & $\bf{0.0057}(31)$$^{*\dagger\ddagger}$\\ 
BNK$^{**}$ & $0.0560(105)$ & $0.0539(89)$ & $0.0531(85)$ & $0.0490(53)$$^{*\dagger\ddagger}$ & $\bf{0.0487}(46)$$^{*\dagger\ddagger}$\\ 
ALR$^{**}$ & $\bf{0.0419}(62)$ & $\bf{0.0419}(69)$ & $0.0420(68)$ & $0.0422(88)$ & $0.0423(91)$\\ 
CA$^{**}$ & $0.0305(117)$ & $0.0294(70)$$^{*}$ & $\bf{0.0293}(66)$$^{*}$ & $\bf{0.0293}(61)$$^{*}$ & $0.0294(64)$$^{*}$\\ 
ELV & $0.0561(84)$ & $0.0561(82)$ & $0.0560(79)$ & $0.0558(69)$ & $\bf{0.0557}(63)$\\ 
CCP & $0.0504(86)$ & $0.0501(78)$ & $0.0501$$\bf{(66)}$ & $\bf{0.0498}$$(68)$ & $0.0504(79)$\\ 
CH$^{**}$ & $\bf{0.1159}(45)$$^{\mathsection\|}$ & $0.1165(63)$$^{\mathsection\|}$ & $0.1167(69)$$^{\|}$ & $0.1192(99)$ & $0.1194(102)$\\ 
CNS & $0.0609(87)$ & $0.0605(80)$ & $0.0603(77)$ & $\bf{0.0599}(64)$ & $0.0602(69)$\\ 
\hline 
\multirow{ 1}{*}{Rank} & \multicolumn{1}{l}{5(63)} & \multicolumn{1}{l}{4(49)} &
\multicolumn{1}{l}{3(45)} & \multicolumn{1}{l}{\bf{1(32)}} & \multicolumn{1}{l}{2(36)} \\ \hline
\end{tabular}
\end{adjustbox}
\egroup
\caption{RMSE for $K_{rel} = 20$} \label{ValidationErrK20}
\end{table}

\begin{table}[!t]
\centering
\bgroup
\def\arraystretch{1}%
\begin{adjustbox}{max width=.8\textwidth}
\begin{tabular}{lllllllllllll}
\hline
\multirow{ 1}{*}{Dataset} & \multicolumn{1}{l}{\textbf{Random}} & \multicolumn{1}{l}{\textbf{RS-K-means++}} & \multicolumn{1}{l}{\textbf{RS-K-medoids++}} & \multicolumn{1}{l}{\textbf{RS-UPGMA}} & \multicolumn{1}{l}{\textbf{RS-maximin}} \\ \hline
AP & $0.0749(89)$ & $0.0704(79)$ & $0.0701(79)$ & $\bf{0.0647}(57)$$^*$ & $0.0682(73)$\\ 
SRV & $0.1072(79)$ & $0.1072(81)$ & $\bf{0.1014}(73)$ & $0.1045(72)$ & $0.1058(72)$\\ 
BC & $0.2679(80)$ & $0.2689(80)$ & $0.2682(77)$ & $0.2666(71)$ & $\bf{0.2662}(62)$\\ 
CHA & $0.0478(94)$ & $\bf{0.0411}(62)$$^*$ & $0.0430(74)$ & $0.0428(70)$ & $0.0436(77)$\\ 
BH$^{**}$ & $0.0843(97)$ & $0.0818(86)$ & $0.0800(73)$ & $\bf{0.0769}(61)$$^*$ & $\bf{0.0769}(61)$$^*$\\ 
FF & $0.0559(80)$ & $0.0566(75)$ & $0.0564(82)$ & $0.0601(79)$ & $\bf{0.0545}(61)$\\ 
STC$^{**}$ & $0.0276(131)$ & $0.0260(97)$$^*$ & $0.0256(75)$$^*$ & $\bf{0.0247}(35)$$^{*\dagger\ddagger}$ & $0.0248(39)$$^{*\dagger\ddagger}$\\ 
S1$^{**}$ & $0.0073(121)$ & $0.0060(78)$ & $0.0059(72)$ & $0.0054(54)$ & $\bf{0.0052}(53)$\\ 
BNK$^{**}$ & $0.0536(105)$ & $0.0527(89)$ & $0.0512(76)$ & $0.0500(58)$$^*$ & $\bf{0.0495}(50)$$^{*\dagger}$\\ 
ALR & $\bf{0.0423}(63)$ & $0.0425(75)$ & $0.0426(79)$ & $0.0426(85)$ & $0.0424(77)$\\ 
CA$^{**}$ & $0.0291(102)$ & $0.0289(74)$ & $0.0289(69)$$^*$ & $\bf{0.0288}(66)$$^*$ & $0.0289(67)$$^*$\\ 
ELV & $0.0572(78)$ & $0.0570(79)$ & $0.0570(81)$ & $\bf{0.0568}(71)$ & $\bf{0.0568}(69)$\\ 
CCP & $0.0491(88)$ & $0.0492(83)$ & $0.0486(67)$ & $\bf{0.0484}(63)$ & $0.0488(76)$\\ 
CH$^{**}$ & $\bf{0.1144}(54)$$^{\mathsection\|}$ & $\bf{0.1144}(61)$$^{\mathsection\|}$ & $0.1145(62)$$^{\mathsection\|}$ & $0.1167(101)$ & $0.1165(99)$\\ 
CNS & $0.0608(79)$ & $0.0605(73)$ & $\bf{0.0603}(71)$ & $0.0605(75)$ & $0.0605(78)$\\ 
\hline 
\multirow{ 1}{*}{Rank} & \multicolumn{1}{l}{5(63)} & \multicolumn{1}{l}{4(49)} &
\multicolumn{1}{l}{3(48)} & \multicolumn{1}{l}{2(34)} & \multicolumn{1}{l}{\bf{1(31)}} \\ \hline
\end{tabular}
\end{adjustbox}
\egroup
\caption{RMSE for $K_{rel}= 40$} \label{ValidationErrK40}
\end{table}
%
%

To demonstrate the differences among the five approaches we examined, we ran only the reference point selection methods for the S1 data, considering 100 reference points (10\%). In Figure \ref{fig:S1RSdists} (Appendix \ref{app:figures}), the smallest 500 pairwise Euclidean distances for the selected 100 reference points for the S1 dataset are plotted in ascending order.
Figure \ref{fig:S1RSdists} also illustrates the differences between the reference point approaches. Overall, Random selection is the worst method and RS-maximin is the best method for identifying separate and input space, covering sets of reference points in a well-balanced manner. Interestingly, the ordering of the methods’ pairwise distance curves is the same as the ordering of the methods’ RMSE performance.

As noted in the results of Section \ref{Sect:fixedK}, variances are not equal for several datasets based on the Brown-Forsythe test. Clustering-based reference point selection gives smaller variances compared to the Random method for a small $K_{rel}$. In Appendix \ref{app:figures}, this is illustrated in Figure \ref{fig:S1RSvariances} and Figure \ref{fig:boxplot} for the S1 dataset. Variance of RMSE for the Random method is 8 times larger compared to the RS-maximin method when $K_{rel} = 5$. When $K_{rel}$ reaches 40, variances are equal.

\subsection{Discussion}

We evaluated four clustering-based methods for the selection of reference points for the MLM. We focused on testing the methods against the Random approach in regression tasks with 15 datasets. An extensive experimental evaluation of the methods shows that the clustering-based methods can improve the performance of the MLM. A good set of reference points is able to cover the data space well. When an optimal number of reference points is desired, RS-UPGMA and RS-maximin are valid choices. With respect to accuracy for a fixed number of reference points $K$, RS-maximin is the best choice for low $K$ values ($K_{rel} = 5, 10$). For higher K values ($K_{rel} = 20, 40$) RS-UPGMA and RS-maximin are the best choices. However, RS-maximin is the most efficient approach, since the computational cost with respect to the number of observations $N$ is linear compared to RS-UPGMA which has a quadratic time complexity with respect to $N$. Together with the LLS method for the second step of the MLM, we obtain, on the whole, a very computationally efficient approach. Note that deterministic reference point selection methods are required to run only once for each dataset in hyperparameter tuning, while, for example, RS-K-medoids++ must be run for each hyperparameter value from the start. Moreover, the deterministic reference point selection methods reduce the MLM model’s space and computational complexity, because they can build the optimal model with smaller sets of reference points.

The conclusion of \cite{pkekalska2006prototype} to favor deterministic strategy agrees with our results on the quality of the deterministic RS-maximin. 
Even though the maximin method is not recommended to be used for the K-means initialization based on the extensive study by \cite{celebi2013comparative}, this study shows that it is a valid method for selecting the reference points in the MLM. This highlights that reference point selection has a different aim than clustering or initialization of a specific clustering method. For example, based on the performed experiments, the maximin method selects points such that extreme points are very valuable if they are not anomalies. Contrarily, in terms of K-means initialization, those points are far from the cluster centers. Hence, they are not optimal choices for clustering initialization.

Finally, the clustering-based methods are less robust for outliers than the Random approach. Therefore, integration with outlier detection or use of a robust approach for input and output distance matrix mapping should be considered for distorted datasets. Based on the experiments, it seems that reference point selection controls the balance between the interpolation and extrapolation of the regression model. Selecting reference points from the boundaries of the data clouds improves extrapolation abilities, but might lead (in rare cases) to worse interpolation in the dense areas, as most likely occurred for the CH dataset.

\section{Conclusion}
\label{concl}

In this paper, we addressed important open questions related to research on the MLM. Based on previous related works, we demonstrated the theory behind the MLM’s interpolation and universal approximation properties by considering the behavior of its two main components, the linear mapping between distance matrices and the multilateration for output estimation. Our results ensure the MLM’s generalization capability and indicate the role of the reference points in the bounded estimation error.

Motivated by our findings, we performed comprehensive computation experiments to evaluate different clustering-based approaches for selecting reference points for the MLM in regression scenarios. In summary, all the methods achieved better performance than standard random selection. The RS-maximin approach was the best choice due to its better generalization capability, compact model size, simplicity, and more efficient computational implementation. In general, our experimental results demonstrate how the utilization of heterogeneous pool of clustering methods with respect to characteristics of an unsupervised problem as a part of supervised method can provide useful insights to the underlying problem.  


In the future, it would be interesting to adapt and evaluate the presented methods for classification tasks as well. Moreover, we could also analyze how such methods are affected if the number of reference points is different for input and output spaces. The latter consideration may modify the reference point selection problem and might result in additional interpretations of the MLM’s generalization capability.






\acks{The work was supported by the Academy of Finland from grants 311877 and 315550. The authors would also like to thank the Cearense Foundation for the Support of Scientific and Technological Development (FUNCAP) for its financial support.}


\newpage

\appendix
\section{Localization Linear System}
\label{ap:LLS}

    Consider $\mathcal{Z}$, a set of known points in $\mathbb{R}^S$. Suppose the existence of $\bm{w} \in \mathbb{R}^S$ is unknown, but whose distances for each $\bm{z}_i \in \mathcal{Z}$, given by $|| \bm{w} - \bm{z}_i ||^2 = d_i^2$, are known. Suppose that we have another point $\bm{r} \in \mathbb{R}^S$, called benchmark-anchor-node (BAN), such that $|| \bm{w} - \bm{r} ||^2 = d_r^2$ and $|| \bm{z}_i - \bm{r} ||^2 = d_{ir}^2$ are also known. Thus, we have:
        
    \begin{align}
        d(\bm{z}_i,\bm{w})^2 = & ||\bm{w} - \bm{z}_i||^2 \nonumber \\
        d_i^2 = & \sum_{j=1}^S (w_j - z_{i,j})^2 \nonumber \\
        d_i^2 = &\sum_{j=1}^S (w_j - r_j + r_j - z_{i,j})^2    \nonumber \\
        d_i^2 = &\sum_{j=1}^S [(w_j - r_j) + (r_j - z_{i,j})]^2 \nonumber \\
        d_i^2 = &\sum_{j=1}^S [(w_j - r_j) - (z_{i,j} - r_j)]^2 \nonumber \\
        d_i^2 = &\sum_{j=1}^S [(w_j - r_j)^2 + (z_{i,j} - r_j)^2 - 2(w_j - r_j)(z_{i,j} - r_j)]  \nonumber \\
        d_i^2 = &\underbrace{\sum_{j=1}^S (w_j - r_j)^2}_{d(\bm{w},\bm{r})^2} + \underbrace{\sum_{j=1}^S(z_{i,j} - r_j)^2}_{d(\bm{r},\bm{z}_i)^2} - 2\sum_{j=1}^S(w_j - r_j)(z_{i,j} - r_j) \nonumber \\
        d_i^2 - d_r^2 - d_{ir}^2 = &- 2\sum_{j=1}^S(w_j - r_j)(z_{i,j} - r_j) \nonumber \\
        \sum_{j=1}^S\underbrace{(w_j - r_j)}_{\theta_j}\underbrace{(z_{i,j} - r_j)}_{A_{ij}} & = \underbrace{\frac{1}{2}[d_r^2 + d_{ir}^2 - d_i^2]}_{b_i} \nonumber \\
        \bm{A} \bm{\theta} = \bm{b} \label{eq:LLS}
    \end{align}
    
        Thus, after solving the system $\bm{A}\bm{\theta} = \bm{b}$, we compute $\bm{w} = \bm{\theta} + \bm{r}$ to recover the position of $\bm{w}$. Note that the BAN $\bm{r}$ can be selected from $\bm{Z}$ and thus satisfy all
        the necessary conditions for the application of the technique.
\section{Figures}
\label{app:figures}

%
%

\begin{figure}[H]
\centering
\includegraphics[width=\textwidth]{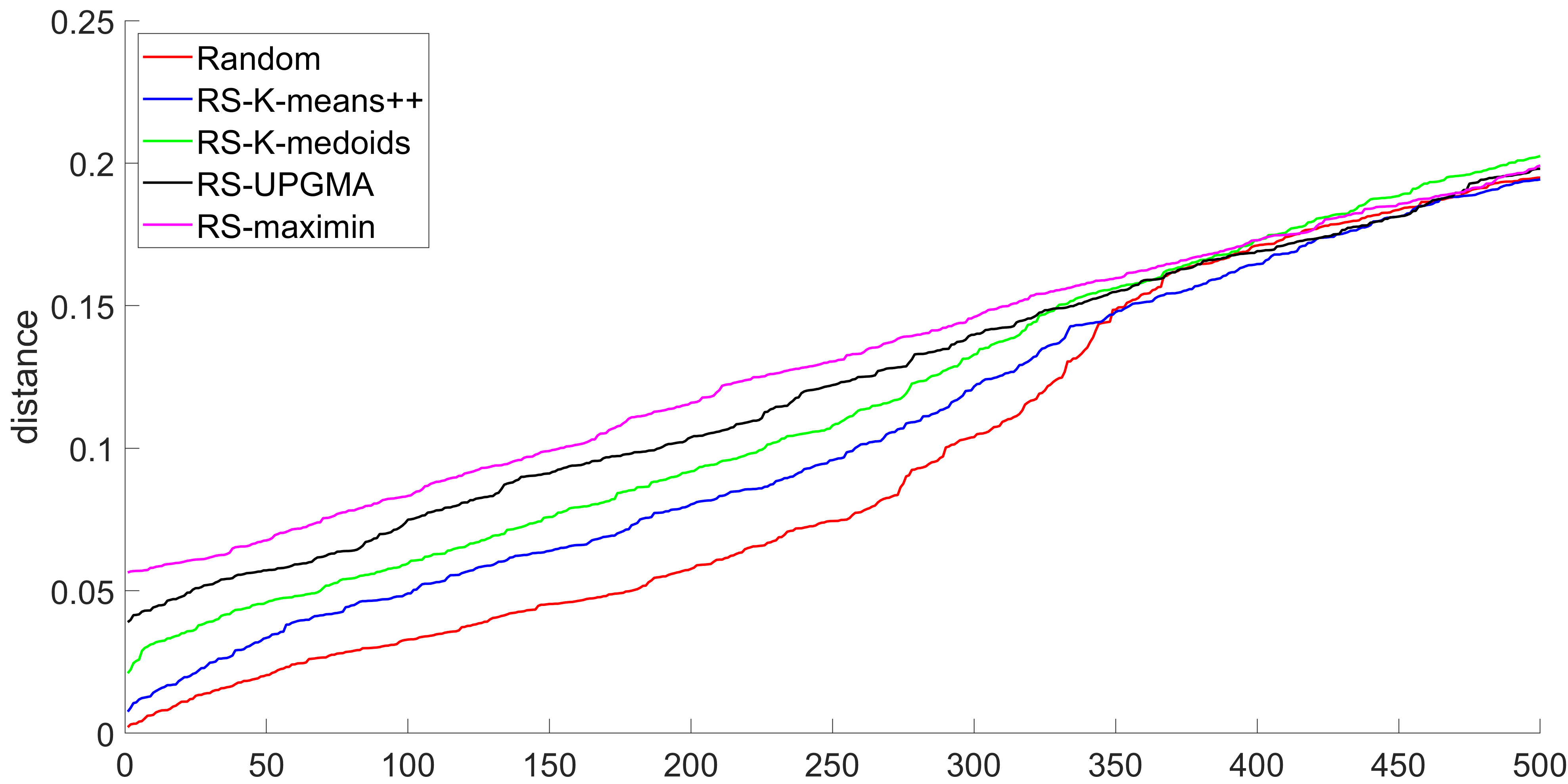}
\caption{The smallest 500 pairwise Euclidean distances for the selected 100 reference points for S1 in ascending order. Clustering-based methods select a set of reference points that are more separeted each other compared to the random approach.}
\label{fig:S1RSdists}
\end{figure}

\begin{figure}[H]
\centering
\includegraphics[width=\textwidth]{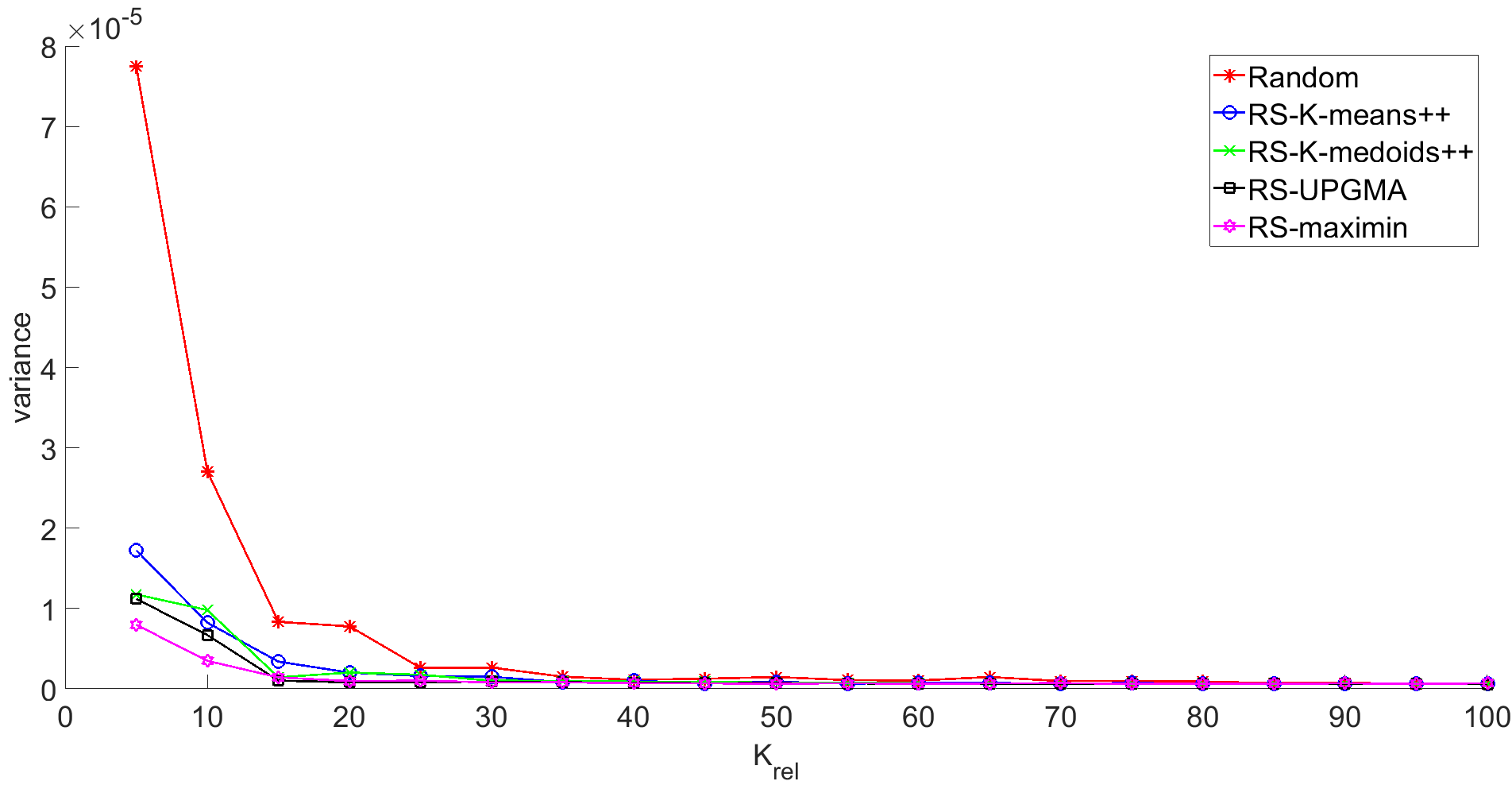}
\caption{Variances of the RMSE test errors for the S1 dataset.}
\label{fig:S1RSvariances}
\end{figure}

\begin{figure}[H]
\centering
\includegraphics[width=\textwidth]{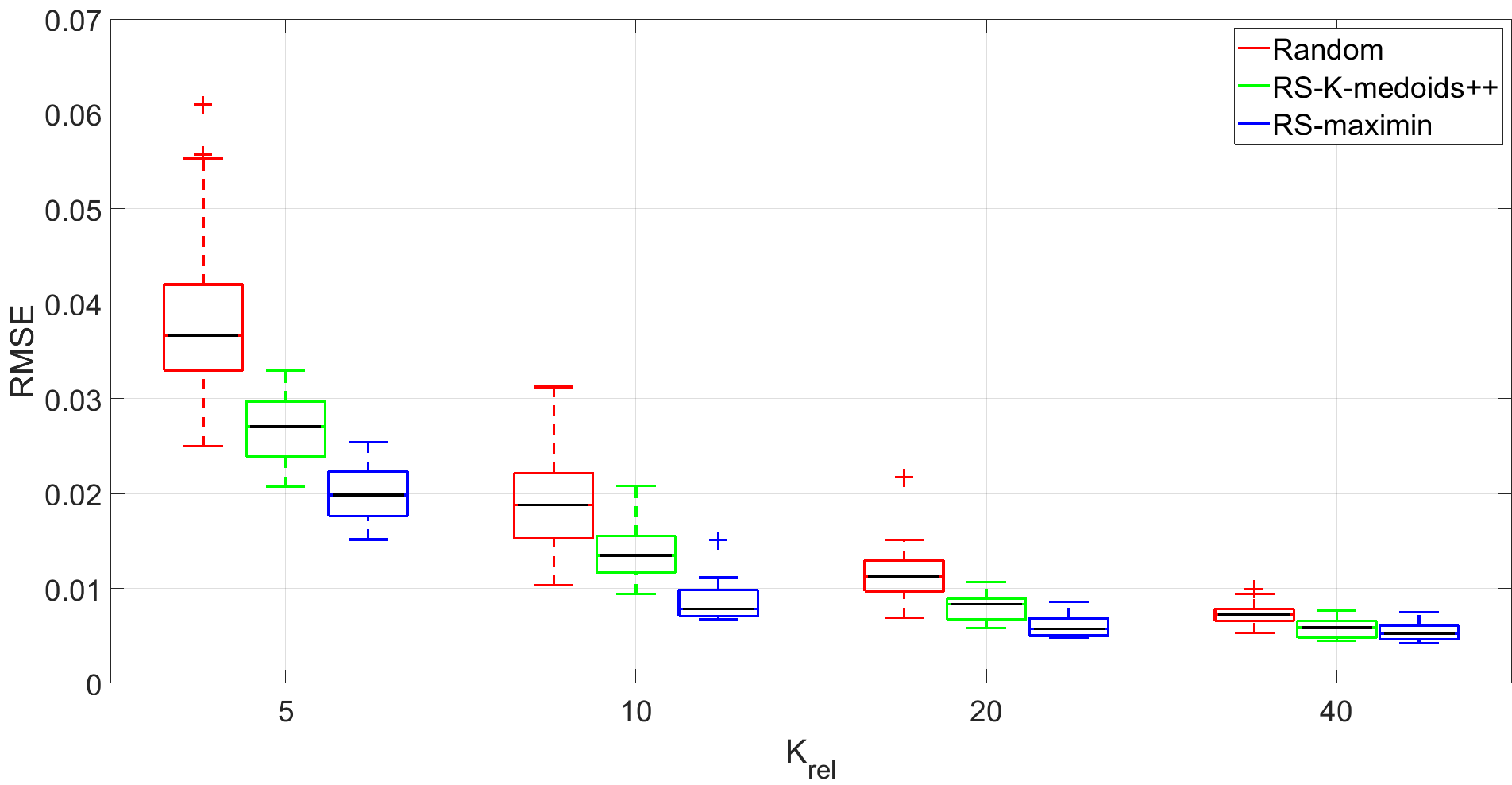}
\caption{Boxplot of the RMSE test errors for the S1 dataset.}
\label{fig:boxplot}
\end{figure}

\clearpage
\bibliography{sample2,AddedRefs}

\end{document}